# Category Trees

Kieran Greer, Distributed Computing Systems, Belfast, UK.
kgreer@distributedcomputingsystems.co.uk.
Version 1.0

*Abstract –* This paper presents a batch classifier that has been improved from the earlier version and fixed a mistake in the earlier paper. Two important changes have been made. Each category is represented by a classifier, where each classifier classifies its own subset of data rows, using batch input values to represent the centroid. The first change is to use the category centroid as the desired category output. When the classifier represents more than one category, it creates a new layer and splits, to represent each category separately in the new layer. The second change therefore, is to allow the classifier to branch to new levels when there is a split in the data, or when some data rows are incorrectly classified. Each layer can therefore branch like a tree - not for distinguishing features, but for distinguishing categories. The paper then suggests further innovations, by adding fixed value ranges through bands, for each column or feature of the input dataset. When considering features, it is shown that some of the data can be classified directly through fixed value ranges, while the rest can be classified using the classifier technique. Tests show that the method can successfully classify a diverse set of benchmark datasets to better than the state-of-the-art. The paper also discusses a biological analogy with neurons and neuron links.

## 1   Introduction

This paper presents a batch classifier that has been improved from the earlier version and fixed a mistake in the earlier paper. Two important changes have been made. Each category is represented by a classifier, where each classifier classifies its own subset of data rows, using batch input values to represent the centroid. The first change is to use the category centroid as the desired category output. This means that for each category, each data column tries to map to a distinct value and it is the summed difference between those values that determines the output error. When the classifier represents more than one category, it creates a new layer and splits, to represent each category separately in the new layer. The second change therefore, is to allow the classifier to branch to new levels when there is a split in the data, or when some data rows are incorrectly classified. Each layer can therefore branch like a tree - not for distinguishing features, but for distinguishing categories. This is different to traditional tree solutions that typically branch on some feature value and so the new classifier is being called Category Trees.

The paper then suggests further innovations, by adding fixed value ranges through bands, for each column or feature of the input dataset. When considering features, it is shown that some of the data can be classified directly through fixed value ranges. An earlier version of the classifier added branches [17] to an oscillating error classification technique [18]. That technique allowed the error update to be independent for each data column, meaning that it could oscillate around the desired output, independent of the other columns. Because that classifier worked off averaged values, it may be the case that some data can be classified directly. The averaged value is simply a single value for a whole range of inputs and so maybe a value band can represent that range as a fixed set of boundaries. It may also be possible to construct these fixed boundaries for single dimensions, representing single features, when much more complex hypercubes are not required. This determination is pre-classifier if you like and would remove categories that can in fact be recognised from distinct features. After categories and related data rows are removed using fixed data bands, the rest can be used to train the classifier. When using for classification, if the input data row does not fall into one of the bands, it can be run through the classifier instead. Tests show that the method can successfully classify a diverse set of benchmark datasets to better than the state-of-the-art. With the idea of these fixed bands, plus the more complex classifiers, the model can be compared with neurons and neuron links.

The rest of this paper is organised as follows: section 2 introduces some related work. Section 3 reviews the earlier work on the classifier, while sections 4 and 5 describe the two improvements. Section 6 describes the possibility of using fixed bands and section 7 re-runs the test set to verify the classifier's accuracy. Finally, section 8 gives a discussion and some conclusions to the work.

## 2     Related Work

This research is based specifically on two earlier papers that introduced an oscillating error classifier [17][18]. The idea of using batch values came from the idea of using the input shape over distinct point values [20]. The input would produce a best fit wave shape between its points, but then it was discovered that the order of the rows could be changed, thereby changing the shape and the averaged shape value would be the same. It was also decided that the averaged wave shape and the average data point values were essentially the same and so batch values were preferred instead. The classifier can be used to classify categorical data, or data rows grouped into categories. It uses averaged values for each category and an oscillating error technique that decides whether to add or subtract the error from each cell value, to minimise the total error. The first paper [18] actually had a mistake in its evaluation, that was corrected by the branching mechanism suggested in [17]. The paper [17] is now replaced by this one, which has extended the work further. The idea of batch processing or averaged values is not new and was used in some of the earlier neural network models, for example [15][23]. The research of this paper also considers classifying each data column, or dimension, separately and this has been looked at previously, usually in relation to nearest neighbour or kNN classifiers [2], for example. The oscillating error technique is a simple rule that introduces the idea of using cellular automata [8][22] as the neural unit, where the small add or subtract decision gives the classifier an added dimension of flexibility. The paper [8] presents a proof that dynamic cellular automata are flexible enough to simulate arbitrary computations, which means algorithms in general. As they describe, this has been put in the context of state machines, where classical algorithms were axiomatized and generalised by Gurevich [22].

Deep Learning Networks [10][28], Decision [21] and Regression Trees have made tremendous advances in a lot of areas, but they are still not universal classifiers. Some recent papers [10] [28][34] show that they can still have problems with these benchmark datasets. The paper [34] compares linear with convex 'hulls' for classifiers and concludes that the linear hull is prone to overfitting and does not work as well. That is interesting, because while this classifier appears to be linear, it performs well with non-linear data. They note that the convex hull is more constrained and each classifier in this new model is definitely bounded. These established classifiers also tend to require a lot more training and configuring before they can produce the better results. The new classifier may also be of interest with respect to biological models of the human brain, which will be elaborated on later, but papers that note the importance of neuron links in the brain for communication processes include [19][36]. Some of the health datasets have been tested, where a summary of the current state there can be found in [26].

## 3     Oscillating Error Classifier Review

This section summarises the earlier oscillating error classifier, used to classify data into categories. Essentially, the classifier works off averaged numerical values and does not incrementally update values for each data row. It would create a classifier for each output category and group all of the input data that belonged to each category. The classifier for that category would then learn to adjust its input, which would be the averaged data row for the data group, to the desired output, which was set to a value like 1 or 0.5, because each category was separate. The idea would be simply to have a weight value, to move the input to the desired category value. The premise for this is the fact that there can only be one weight value for all of the inputs and so learning the averaged value looks reasonable. The oscillating error method added or subtracted the difference between the actual and the desired value from each data column value separately. So, for example, the difference could be subtracted from the averaged value in column 1, but added to the averaged value in column 2. This was repeated until a minimum error or a maximum number of iterations was achieved. The data was also normalised, to be in the range 0 to 1, so that each adjustment was equal. This oscillating error process was still not accurate enough, where lots of data rows would still be incorrectly classified. Each classifier therefore branched to a new level when it was associated with data rows from more than 1 category, see section 5. Any new level would create a new set of classifiers, one for each associated category and repeat the process in exactly the same way as for the parent level. This new level however would have a less complicated problem to solve, because it would use only a subset of the whole dataset, related to the parent classifier only. It is probably also the case that the oscillating error technique is not required for a model that uses a separate classifier for each category, and a single adjustment from the averaged value to the desired output value can be used instead. This is the basis for section 4, where the adjustment can still be an addition or subtraction however, and made independently for each column.

When using the classifier then: if it had branches, they were always passed the input data and asked to return their evaluation. If there was only 1 classifier with no branches, it would return its own result, which was an error measurement of how far the input data row was to the category value, after the weight adjustment. Therefore, each base classifier would return some category evaluation and error result and the classifier with the smallest error would be selected as the best match. Logically this could lead to a classifier learning a single data row and that might be expected to be 100% accurate. This is not quite the case and overfitting is an obvious problem from it. Overlapping regions in the data values probably lead to some confusion as well.

## 4   Map to the Centroid Value

The first improvement to the classifier is to map directly to the centroid value for the category dataset. Each category is assigned a separate classifier and the desired category output value was originally set to something artificial. This might have been 1 or 0.5, or in some cases a graded value through all of the categories was preferred. The value was not as important as the weight adjustment to it and so the classifier can in fact be simplified from that model by using the category centroid for the desired category output and subsequent weight adjustment. This also helps to add non-linearity, through a curved output line, for example. Each new input data row can be compared to the centroid directly and the total absolute difference in the row values taken to be the error. The following example explains how it works: Consider a dataset with 2 categories A and B. These datasets have the following rows, shown in Figure 1, assigned to them. A new data row has the values '4, 3, 2' and comparing to the centroids gives the error values, also shown in Figure 1. Therefore, the new data row belongs to category A.

| Category A | Category B |
|---|---|
| Data row: 1, 2, 1 | Data row: 5, 6, 5 |
| Data row: 1, 2, 3 | Data row: 5, 6, 7 |
| Data row: 3, 2, 1 | Data row: 7, 6, 5 |
| Centroid row: 1.66, 2, 1.66 | Centroid row: 5.66, 6, 5.66 |
| Difference = 2.34+1+0.34 = 3.68. | Difference = 1.66+3+3.66 = 8.32 |

Figure 1. Data rows, centroid and error difference with data row '4, 3, 2', for 2 categories.

## 5   Extending the Classifier with Branching

This section is repeated from the paper [17] and describes the second improvement. The dataset has therefore been split into groups, and there are *x* classifiers in the first level, one for each category. Each classifier is assigned its centroid as the desired output. The whole train dataset is then passed through all of the classifiers and each produces an error for each data row, as described in Figure 1. After this test phase, there is a list of data rows that each classifier has produced the closest match to, with relation to its own category. Most of the rows would be for the correct category group, but some would be for other classifier groups. The branching extension therefore adds a new level to the classifier, to refine it with respect to the incorrectly classified data rows. The schematic of Figure 2 shows the classification process, where a new layer has been added to classifier A, so that it can correctly re-classify the category A and B sub-groups that belong to it. The second level uses a subset of the whole dataset that is specifically only the data rows that the first level classified as closest. For the classifier's own category, this is probably almost the same as for the first level. For any other categories, there are new classifiers in the layer to reclassify those correctly.

### 5.1   Mathematical Proof

There is some mathematical justification for why centroid values can be used for the output category values. Because averaged values are used, the problem is to map as closely as possible to these and that is a bit like finding a best fit line. In this case, the best fit line is known and would be the centroid values. Therefore, it makes sense to try to map to this and make it the desired output value as well. The classifier then needs to converge when data rows are incorrectly classified. When training, if the classifier moves to its next layer, it only needs to consider the dataset rows related with its current layer. So, when these are adjusted for incorrect categories, the consideration is for that subset only. This is a fairly basic argument for why the classifier should work.

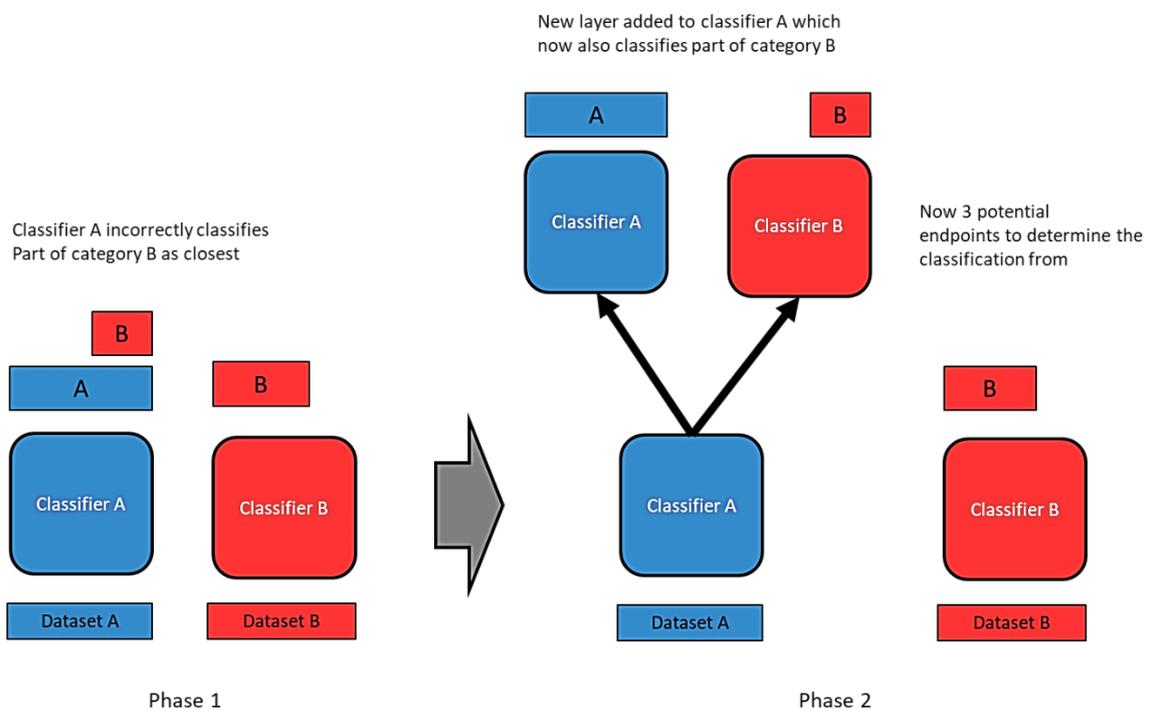

Figure 2. Schematic of the classifier in action. Phase 1 realises that classifier A also classifies part of category B better. Phase 2 then adds a new layer to classifier A, to re-classify this subset only.

## 6   Extending the Classifier with Fixed Bands

If the classifier deals with averaged values, then it does not try to learn too much outside of that and so one question might be if ranges of input values can be classified directly. It would certainly be the case if the data was linearly separable, because the separating line would allow a clear distinction to be determined, but there is still a problem when data from different categories overlap. If trying to separate the input data then, multi-dimensional hypercubes would be the first choice, but this looks like a very difficult model to implement. Therefore, another option might be to try to separate on each dimension, or feature only and the test results show that this can be quite successful. The process is as follows: The data can be read, one column at a time, as it is organised in the data file. The category that each row belongs to is also retrieved and if there is a change in category, the previous set of values can be placed into a band. The only problem is when categories overlap in a single dimension, which would mean that they have the same value, when the band then continues to the next value and category change. For example, consider the following values for a column and related categories, shown in Figure 3.

| Column Value | Category |
|---|---|
| 0.1 | A |
| 0.2 | A |
| 0.3 | A |
| 0.4 | B |
| 0.5 | B |
| 0.5 | C |
| 0.6 | C |

Figure 3. Example of Data column values with related categories, placed into bands: Band 1 - 0.1 to 0.3 and Band2 - 0.4 to 0.6.

The program would therefore firstly sort the data column values into order. It would then read down the column until there is a change in the category. In this case, the first change is at the value 0.4. The value range 0.1 to 0.3 all belongs to a single category and so a band can be made from that. Then process continues and the next break would be at the value 0.5, but there is an overlap with this value as both categories B and C use it. Therefore, the process must continue to the end value, when both of those categories are placed in a single value range of 0.4 to 0.6.

The process is repeated for each column of to produce a set of bands for the column. It is also important to link the bands from one dimension to the next depending on the exact values in each data row. As an example, if there are 3 columns in a dataset and each column has 5 bands; then if a data row relates to bands 1, 2 and 4, these bands will have links added between them. Then the band 1 relating to column or dimension 1 can only move to band 2 in the next dimension, and so on. Figure 4 shows the bands and links created for the Iris Plants dataset [11].

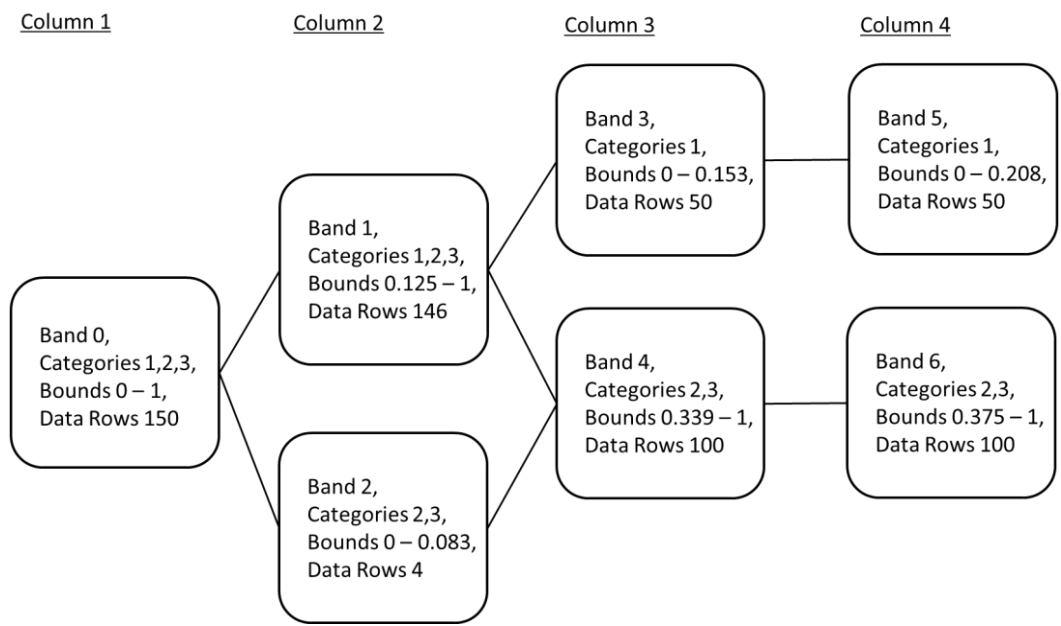

Figure 4. Bands created for the Iris Plants dataset [11].

When presented with a new data row to evaluate, the procedure traces through the band links, to check if any represent a single category only. If that is the case, then the fixed band ranges can be used to classify the input data directly. Because of the overlap however, there are lots of cases where a band represents more than 1 category. It would be interesting to train the classifiers for those cases only, but for a first test, the classifier system with branching was also trained and used if the bands did not return a result. The bands can help a bit here however, because if they can classify and data rows directly, those rows do not need to be considered by the classifiers and so this was also checked for and those data rows removed from the training dataset. The bands can therefore make a significant contribution to the classification process. For example, Figure 4 shows bands, where category 1 is linearly separable and can be identified completely from using the band ranges. Categories 2 and 3 would need a classifier to be separated. Also, for category 1, the classification is clear from column 3 and so column 4 is probably not required. In effect, column 3 provides a unique feature for category 1.

## 7 Test Results

This paper repeats the set of tests carried out in the earlier paper [18], to verify the classifier accuracy and includes some new datasets. A test program has been written in the C# .Net language. It can read in a data file, normalise it, generate the classifier from it and measure how many categories it subsequently evaluates correctly. Each output category is now represented by a separate classifier and the centroid vector. Two types of result were measured. The first was an average error for each row in the dataset, after the classifier was trained, calculated as the average difference between the input vector and the output centroid vector. This is only a minor result and the second more important measurement was how many categories were correctly classified. For these tests, the error margin used in the original tests was not required, where the errors of each output classifier could be compared directly and the smallest one selected.

### 7.1 Benchmark Datasets with Train Versions Only

The classifier was first tested on several datasets from the UCI Machine Learning Repository [39]. These included the Wine Recognition [12], Iris Plants [11] and the Zoo [41] databases. Wine Recognition and Iris Plants have 3 categories, while the Zoo database has 7. These do not have a separate training dataset and are benchmark tests for classifiers. In fact, the classifier trains in a single step and so it does not require a stopping criterion, but it could be prone to over-training if layers are continually added. For the Wine dataset, the UCI [39] web page states that the classes are separable, but only RDA [14] has achieved 100% correct classification and results have used the leave-one-out technique. Other datasets included the Abalone shellfish dataset [1], the Hayes-Roth concept learning dataset [24] and the BUPA Liver dataset [32]. Then the Cleveland Heart Disease [9] and the Breast Cancer [25] datasets were tested as well. While the Heart Disease dataset was originally tested for presence (cats 1-4) or absence (cat 0), this test matched with the output category exactly.

Another web site [5] lists other datasets, where tested here is the Sonar [16][40] and Wheat Seeds [6] datasets, with previous benchmark results of 100% and 92% respectively. The Car [3][35] and Wine Quality [7] datasets were also tested. As shown in Table 1, the new classifier produces a remarkable result of 100% accuracy over all of these datasets. The column 'Other Best %' lists a result found by some other researchers. The final column indicates if feature bands could be used and if they were better. For these tests, using bands did not affect the result and improved the result in only the Hayes-Roth case.

| Dataset | Av Error | Number Correct | % Correct | Other Best % | Use Bands |
|---|---|---|---|---|---|
| Wine | 0.1 | 133 from 133 | 100 | 100 | Either |
| Iris | 0.07 | 150 from 150 | 100 | 97 | Either |
| Zoo | 0.1 | 101 from 101 | 100 | 94.5 | Either |
| Abalone | 0.01 | 4177 from 4177 | 100 | 73 | Either |
| Hayes-Roth | 0 | 132 from 132 | 100 | 50 | Yes |
| Liver | 0.08 | 345 from 345 | 100 | 74 | Either |
| Cleveland | 0.15 | 303 from 303 | 100 | 77 | Either |
| Breast | 0.08 | 569 from 569 | 100 | 98.5 | Either |
| Sonar | 0.15 | 208 from 208 | 100 | 100 | Either |
| Wheat | 0.09 | 210 from 210 | 100 | 92 | Either |
| Car | 0.28 | 1728 from 1728 | 100 | 97 | Either |
| Wine Quality | 0.06 | 1599 from 1599 | 100 | 89 | Either |

Table 1. Classifier Test results. Average output error and number of correct classifications. All datasets points normalised to be in the range 0 to 1.

### 7.2   Separate Train and Test Datasets

There is an important question about generalisation properties when averaged values and bands are used, and the fact that batch data rows can reduce to a number of 1. A slightly better test would therefore be to have different train and test datasets and this section gives the results for those tests, shown in Table 2. The training time would be instantaneous for something like the Iris Plants dataset, but for the Letter Recognition dataset in this section, it could take longer. But the system is very low on resource usage and the process is completely deterministic, meaning that it should give exactly the same result each time. There is no fine tuning either.

Another set of test results, using separate train and test datasets, was as follows: The first set of datasets were User Modelling [27], Bank Notes [33], SPECT images heart classification [30], Letter recognition [13], the first Monks dataset [38] and Solar flares [4][31]. Then two other datasets were artificially split into a train and test set. The Pima Indians Diabetes dataset [29] was split into a train set of 400 rows and a test set of 368 rows. The Ionosphere dataset [37] was divided into a train set of 201 rows and a test set of 150 rows. Both of these performed equally well on the whole dataset. It can be seen however that for these tests the classifiers preferred not to use fixed bands and so the generalising properties of the bands is less than for the classifier itself.

| Dataset | Av Error | Number Correct | % Correct | Other Best % | Use Bands |
|---|---|---|---|---|---|
| **UM** | 0.17 | 144 from 145 | 99.9 | 98 | No |
| **Bank** | 0.15 | 100 from 100 | 100 | 61 | No |
| **SPECT** | 0.14 | 187 from 187 | 100 | 84 | No |
| **Letters** | 0.07 | 3623 from 4000 | 90 | 82 | No |
| **Monks-1** | 0.35 | 432 from 432 | 100 | 100 | Either |
| **Solar** | 0.05 | 1017 from 1066 | 95 | 84 | Yes |
| **Diabetes** | 0.12 | 368 from 368 | 100 | 77 | No |
| **Ionosphere** | 0.15 | 150 from 150 | 100 | 96 | Either |

Table 2. Classifier Test results. The same criteria as for Table 1, but a separate test dataset to the train dataset.

### 7.3   Test Conclusions

The new version that includes bands is certainly worth considering, even if the earlier version can produce better results in more cases. Both versions are very easy to train and use. The size of the whole structure is quite large, but it is also very simple and the same each time. For example, for the Abalone dataset, 2700 classifiers were created and the Letters dataset produced 12600 classifiers, where most of that would be branching to next levels. In real time, it might then have to test the input across that number of classifiers as well, which is quite a lot for a relatively small dataset. While using bands should reduce the number of data rows that a classifier needs to learn, it typically resulted in even more classifiers a well, and so there may be a coherence factor over the dataset. They were also shown not to be as useful on previously unseen data and so their generalisation properties are not as good. However, for a couple of datasets, they provided better results. While each classifier solves only a small part of the larger problem, it is not the case that each classifier has been given a few rows of data to classify. The system has generated the classifiers and row sets for itself. It should also be possible to update the system dynamically.

### 7.3.1   Dataset Normalisation

The results in Table 1 are very impressive, but maybe not the whole story. The Wine Quality dataset [7] illustrates another point. For one test, it was divided into a train set of 1100 data rows and a separate test set of 499 rows. Normalising the data was then done in one of two ways. The first way would be to normalise over each dataset separately. This would potentially create different minimum and maximum values for each dataset. The other way would be to normalise over both datasets together and use the same minimum and maximum value to normalise either with. This could potentially change the result of the classification for a previously unseen dataset. In the case of the wine quality data, if the train and test sets were normalised separately, the test set would be recognised at only 86.5% accuracy (431 correct from 499). If the datasets were normalised together, then the test set was recognised at 90% accuracy (451 correct from 499), which is on a par with the other selected best value. This is therefore another factor to consider and could lead to better results if the correct option is chosen.

## 8   Conclusions

This paper has extended the work reported in [17][18], with two improvements. These are to use the centroid as the matching criterion for the output category and to branch the classifier when some data rows are incorrectly classified. As the structure is now a tree, the classifier can be called Category Trees. The idea of fixed bands is also introduced, where these bands can be used to classify some input data directly, simply by using value ranges.

The bands and the classifiers are currently trained separately and are not linked-up, but the bands can remove some data rows for the training of the classifiers. The test results show that it can out-perform a lot of other classifiers and is probably easier to use. The exact nature of the classifier is now a lot clearer. It simply maps to the averaged values through a linear adjustment, but non-linearities can benefit from the distributed and unrelated nature of the stored weight values. These non-linearities however, can be built-up systematically and holistically, which explains why it can classify something like the Abalone shellfish dataset [1] better. It is not helped by other methods, for overfitting for example, but the results are such that it would take something unusual to improve it further.

### 8.1  Biological Discussion

Comparing the classifier with something like PCA, feature selection and biological systems can lead to some logical conclusions. If a data object has a unique feature (data column) then a data band can use that to classify it directly. Maybe in effect, it could be passed down a unique channel relating to that band. If the data object does not have a unique feature, then the current system takes all of the features together and compares that with an averaged value of the features present in each category. The data object is then allocated the category that it is a closest match to. The comparison with neurons and links between them is clear. There is biological evidence that the links between neurons play an important role in the actual signal interpretation and understanding in the human brain [36]. The bands would therefore be an analogy to the neuron links. If the signal fits inside of the band boundaries, it can be classified as whatever the band represents without a neuron interpreting it further. If there is any discrepancy, then a classifier is required to sort that out and so this is analogous to a neuron being created to process a more mixed signal. If the neuron behaves like a filter, then the process might be to convert the mixed signal back into more singular parts again.